# Fault Classification using Pseudomodal Energies and Neuro-fuzzy modelling

___

TSHILIDZI MARWALA, THANDO TETTEY AND SNEHASHISH CHAKRAVERTY

**ABSTRACT:**

   This paper presents a fault classification method which makes use of a Takagi-Sugeno neuro-fuzzy model and Pseudomodal energies calculated from the vibration signals of cylindrical shells. The calculation of Pseudomodal Energies, for the purposes of condition monitoring, has previously been found to be an accurate method of extracting features from vibration signals. This calculation is therefore used to extract features from vibration signals obtained from a diverse population of cylindrical shells. Some of the cylinders in the population have faults in different substructures. The pseudomodal energies calculated from the vibration signals are then used as inputs to a neuro-fuzzy model. A leave-one-out cross-validation process is used to test the performance of the model. It is found that the neuro-fuzzy model is able to classify faults with an accuracy of 91.62%, which is higher than the previously used multilayer perceptron.

## INTRODUCTION

   The process of monitoring and identifying faults in structures is of great importance in aerospace, civil and mechanical engineering. Aircraft operators must be sure that aircraft are free from cracks.  Bridges and buildings nearing the end of their useful life must be assessed for load-bearing capacities. Cracks in turbine blades lead to catastrophic failure of aero-engines and must be detected early.   Many techniques have been employed in the past to locate and identify faults.   Some of these are visual (e.g. dye penetrant methods) and others use sensors to detect local faults (e.g. acoustics, magnetic field, eddy current, radiographs and thermal fields). These methods are time consuming and cannot indicate that a structure is fault-free without testing the entire structure in minute detail. Furthermore, if a fault is buried deep within the structure it may not be visible or detectable by these localised techniques. The need to detect faults in complicated structures has led to the development of global methods which are able to utilise changes in the vibration characteristics of the structure as a basis of fault detection [1].
   There are four main methods by which vibration data may be represented: time, modal, frequency and time-frequency domains.   Raw data is obtained from measurement made in the time domain.   From the time domain, Fourier transform techniques may then be used to transform data into the frequency domain.

___

University of the Witwatersrand, {t.tettey, t.marwala}@ee.wits.ac.za. Central Building Research Institute, sne_chak@yahoo.com

From the frequency domain data, and sometimes directly from the time domain, the modal properties may be extracted. All of these domains theoretically contain similar information but in reality this is not necessarily the case. Because the time domain data are relatively difficult to interpret, they have not been used extensively for fault identification, and for this reason, the modal properties have been widely considered. In this paper we use the pseudomodal, defined from the Frequency Response Function (FRF), together with a Takagi-Sugeno (TS) Neuro-fuzzy model to classify faults in a population of cylindrical shells.

## BACKGROUND

**Pseudomodal Energies**

In this work, Pseudomodal Energies are used for the classification of faults in cylinders. Pseudomodal energies have been found to allow for better classification of faults when compared to modal properties. Pseudomodal energy is defined as the integral of the frequency response function (FRF) over various frequency bandwidths [2]. The FRF is defined as the ratio of the Fourier transformed response to the Fourier transformed force. The pseudomodal energies are therefore the intergral of the real and imaginary parts of the FRFs over various frequency ranges that bracket the natural frequencies.

On one hand, receptance expression of the FRF is defined as the ratio of the frequency response of displacement to the frequency response of force. On the other hand, inertance expression of the FRF is defined as the ratio of the frequency response of acceleration to the frequency response of force. Similarly, the pseudomodal energies can be expressed in terms of the receptance and inertance. The commonly used techniques of collecting vibration data involve measuring the acceleration response and therefore it is more useful to calculate the inertance pseudomodal energies. The inertance pseudomodal energy is derived by integrating the inertance FRF written in terms of the modal properties by using the modal summation equation as follows

$$\text{IME}_{kl}^q = \int_{a_q}^{b_q} \sum_{i=1}^{N} \frac{-\omega^2 \varphi_k^i \varphi_l^i}{-\omega^2 + 2\zeta_i \omega_i \omega j + \omega} d\omega . \tag{1}$$

where $a_q$ and $b_q$ represent, respectively, the lower and the upper frequency bounds for the $q$th pseudomodal energy calculated from the FRF due to excitation at $k$ and measurement at $l$. $N$ is the number of mode and $\zeta_i$ is the damping ratio of mode $i$. For a detailed derivation of this equation consult [3]. Assuming the damping is low, Eq (1) becomes [2]:

$$IME_{kl}^q \approx \sum_{i=1}^{N} \left\{ \varphi_k^i \varphi_l^i (b_q - a_q) - \omega_i \varphi_k^i \varphi_l^i j \left[ \arctan\left(\frac{-\zeta_i \omega_i - jb_q}{\omega_i}\right) - \arctan\left(\frac{-\zeta_i \omega_i - ja_q}{\omega_i}\right) \right] \right\} \tag{2}$$

The advantage in using IMEs over the use of the modal properties is that all the modes in the structure are taken into account as opposed to using the modal properties, which are limited by the number of modes identified; and integrating the FRFs to obtain the pseudomodal energies smoothes out the zero-mean noise present in the FRFs.

**Neuro-fuzzy modelling**

A fuzzy inference system is a model that takes a fuzzy set as an input and performs a composition to arrive at the output based on the concepts of fuzzy set theory, fuzzy *if-then* rules and fuzzy reasoning [4]. Simply put, the Fuzzy inference procedure involves: the fuzzification of the input variables, evaluation of rules, aggregation of the rule outputs and finally the defuzzification of the result. There are two popular types of fuzzy models: the Mamdani model and the Takagi-Sugeno model. The Takagi-Sugeno model is popular when it comes to data-driven identification and is used in this study. In this model the antecedent part of the rule is a fuzzy proposition and the consequent is an affine linear function of the input variables as shown in (3) [5].

$$R_i : \text{If } x \text{ is } A_i(x) \text{ then } y_i = a_i^T x + b_i, [w_i] \qquad (3)$$

Where $a_i$ is the consequence parameter vector, $b_i$ is a scalar offset and $i = 1, 2..., K$.

The symbol *K* is the number of fuzzy rules in the model and $w_i \in [0,1]$ is the weight of the rule. The antecedent propositions in the model describe the fuzzy regions in the input space in which the consequent functions are valid and can be stated in the following conjunctive form:

$$R_i : \text{If } x_1 \text{ is } A_{i,1}(x_1) \text{ and ... and } x_n \text{ is } A_{i,n}(x_n) \text{ then } \hat{y} = a_i^T x + b_i, [w_i]. \qquad (4)$$

The degree of fulfilment of the *i*th rule is calculated as the product of the individual membership degrees and the rule's weight:

$$\beta_i(x) = w_i A_i(x) = w_i \prod_{j=1}^{n} A_{i,j}(x_j) \qquad (5)$$

The output *y* is then computed by taking a weighted average of the individual rules' contributions as shown below:

$$\hat{y} = \frac{\sum_{i=1}^{K} \beta_i(x) y_i}{\sum_{i=1}^{K} \beta_i(x)} = \frac{\sum_{i=1}^{K} \beta_i(x)(a_i^T x + b_i)}{\sum_{i=1}^{K} \beta_i(x)} \qquad (6)$$

Where $\beta_i(x)$ is the degree of fulfilment of the *i*th rule. The parameters $a_i$ are then approximate models of the considered nonlinear system.

Fuzzy rule-based systems with learning ability, also known as neuro-fuzzy networks [6], will be considered in this work. This system will be referred to as a neuro-fuzzy system (model) from here onwards. There are two approaches to training neuro-fuzzy models [7]:

1. Fuzzy rules may be extracted from expert knowledge and used to create an initial model. The parameters of the model can then be fine tuned using data collected from the operational system being modelled.
2. The number of rules can be determined from collected numerical data using a model selection technique. The parameters of the model are also optimised using the existing data.

The second approach is used in this study as there is no expert knowledge which will allow us to create an adequate initial model.

**EXPERIMENTAL SETUP**

**Data gathering and pre-processing**

The data in this work is obtained by performing an experiment on a population of cylinders, which are supported by inserting a sponge rested on a bubble-wrap, to simulate a 'free-free' environment. This setup is illustrated in figure 1 below. The sponge is inserted inside the cylinders to control boundary conditions. This will be further discussed below. Conventionally, a 'free-free' environment is achieved by suspending a structure usually with light elastic bands. A 'free-free' environment is implemented so that rigid body modes, which do not exhibit bending or flexing, can be identified. These modes occur at frequency of 0Hz and they can be used to calculate the mass and inertia properties. In the present study, we are not interested in the rigid body modes. Testing the cylinders suspended is approximately the same as testing it while resting on a bubble-wrap, because the frequency of cylinder-on-wrap is below 100Hz. The first natural frequency of cylinders being analysed is over 300Hz and this value is several order of magnitudes above the natural frequency of a cylinder on a bubble-wrap. Therefore the cylinder on the wrap is effectively decoupled from the ground. It should be noted that the use of a bubble-wrap adds some damping to the structure but the damping added is found to be small enough for the modes to be easily identified. The impulse hammer test is then performed on each of the 20 steel seam-welded cylindrical shells. The impulse is applied at 19 different locations as indicated in Figure 1: 9 on the upper half of the cylinder and 10 on the lower half of the cylinder. The sponge is inserted inside the cylinder to control boundary conditions by rotating it every time a measurement is taken. The top impulse positions are located 25mm from the top edge and the bottom impulse positions are located 25mm from the bottom edge of the cylinder. The angle between two adjacent impulse positions is 36$^o$.

For one cylinder the first type of fault is a zero-fault scenario. This type of fault is given the identity [0 0 0], indicating that there are no faults in any of the three substructures. The second type of fault is a one-fault-scenario, where a hole may be located in any of the three substructures. Three possible one-fault-scenarios are [1 0 0], [0 1 0], and [0 0 1] indicating one hole in substructures 1, 2 or 3 respectively. The third type of fault is a two-fault scenario, where one hole is located in two of the three substructures. Three possible two-fault-scenarios are [1 1 0], [1 0 1], and [0 1 1]. The final type of fault is a three-fault-scenario, where a hole is located in all three substructures, and the identity of this fault is [1 1 1]. There are 8 different types of fault-cases considered (including [0 0 0]).

Each cylinder is measured three times under different boundary conditions by changing the orientation of a rectangular sponge inserted inside the cylinder. The number of sets of measurements taken for undamaged population is 60 (20 cylinders X 3 for different boundary conditions).

The impulse and response data are processed using the Fast Fourier Transform (FFT) to convert the time domain impulse history and response data into the frequency domain. The data in the frequency domain are used to calculate the FRFs. From the FRFs, the modal properties are extracted using modal analysis and the pseudo modal energies are calculated using the integrals under the peaks for a given frequency bandwidth using the trapezoidal technique. The frequency spacing of the FRFs is 1.22Hz. When the pseudo modal energies are calculated, frequency ranges spanning over 6% of the natural frequencies are chosen. These bandwidths are as follows in Hz: 393-418, 418-443, 536-570, 1110-1180, 1183-1254, 1355-1440, 1450-1538, 2146-2280, 2300-2440, 2250-2401, 2500-2656, 3140-3340, 3350-3565, 3800-4039, and 4200-4458. The guidelines outlined in [2] are taken into consideration when choosing these frequency ranges. These guidelines state that the frequency bandwidth must be: (1) sufficiently narrow to capture the resonance behavior, (2) sufficiently wide to capture the smoothing out of zero-mean noise, and (3) must not include the regions of the anti-resonance, which are generally noisy. The pseudo modal energies are used to train the pseudo-modal-energy TS neuro-fuzzy model. The numbers of pseudo-modal-energies identified are 646 (corresponding to 17 natural frequencies X 19 measured mode-shape-co-ordinates X 2 for real and imaginary parts of the pseudo modal energy). The Statistical Overlap Factor (SOF) and the Principal Component Analysis (PCA) are used to reduce the dimension of the input data from 646X167 pseudo-modal-energies to 10X167 for both these data types.

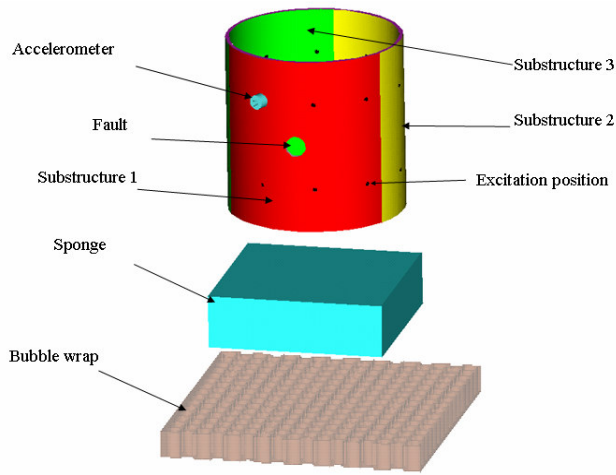

Figure 1: A schematic diagram illustrating the set up of the experiment.

**Neuro-fuzzy model optimisation**

For the optimization of the neuro-fuzzy model a 10-fold cross validation method is used for model selection. Once the appropriate model has been selected, the training data is then used to train a neuro-fuzzy model with the correct model parameters. A first order TS neuro-fuzzy model with a Gaussian membership function has been implemented. The optimum number of rules in the model has been determined by evaluating models with between one and ten fuzzy rules. The optimum number of rules is found to be four as it gives a low prediction error together with a small standard deviation. The remaining parameters of the neuro-fuzzy model are optimised using a combination of the least squares and gradient descent methods.

**Threshold determination**

For a given input set, the neuro-fuzzy model gives an output of three decision values [x, y, z]. A correct classification must give the correct values of x, y and z i.e. it must correctly predict all the faults in one cylinder. The neuro-fuzzy model gives ouput values in the range [0, 1], and it is expected that the decision point will be around the halfway mark i.e. 0.5. In this experiment, two separate methods of determining the output have been tested. One simply assumes a decision point of 0.5 and the other method finds a decision point which minimises the error on the training set of 167. Individual thresholds are evaluated for each of three fault areas on the cylindrical shell. The selected threshold is the one that yields the maximum accuracy as defined by:

$$Acc = pos \cdot tpr + neg \cdot (1 - fpr) = \frac{tpr + c(1 - fpr)}{c + 1} \qquad (7)$$

Where tpr is the true positive rate, also known as the sensitivity given by:

$$tpr = TP/(TP + FN) \qquad (8)$$

In Eq 7, fpr is defined as the false positive rate also known as the specificity given by:

$$fpr = FP/(FP + TN) \qquad (9)$$

TP, FN, FP and TN are all obtained from a confusion matrix and are defined as true positive, false negative, false positive and true negative, respectively. Parameter $c$ is the relative importance of negatives to positives. In this study the fault cases and non-fault cases have been given equal importance in classification, meaning $c$ has been assigned a value of 1. The results obtained from using both the threshold selection techniques are given in the next section. It should be noted that a correct classification is one in which the classifier correctly predicts the condition of all the three substructures of the cylinder.

**Generalisation performance**

One of the problems experienced in machine learning is the assessment of the generalisation capabilities of a model. The $K$-fold cross-validation method has been shown to be an improved measure of performance over the holdout method, which divides the dataset into training and testing set [8]. With $K$-fold cross-validation, the dataset is divided into $K$ approximately equal sets. The holdout method is then performed $K$ times, where each time one of the unique $K$ sets are held back as a testing set and the model is optimised using the combined, remaining $K$-1 sets. The generalisation estimate is then the average error of the model over all the $K$ sets. Leave-one-out cross-validation is $K$-fold cross validation taken to its extreme. In this case $K$ is equal to $N$, the number of instances in the given dataset. The cross-validation technique, though computationally expensive, is useful especially in the cases where modelling data is very limited. Moreover, it has been shown that not only is the leave-one-out cross validation generalisation estimate is a better measure, the worst-case error of this estimate is not much worse than that of the training error estimate [8]. In our work, the number of data points we have is limited to 167. We therefore use the leave-one-out cross-validation method to measure the performance of the TS neuro-fuzzy model.

**RESULTS AND DISCUSSIONS**

The classification result is assessed using the leave-one-out cross validation process described above. The performance of the TS neuro-fuzzy model is illustrated are shown in Table 1 below.

Table 1: The table shows the classification results that are obtained when using the neuro-fuzzy model

| Method | Misclassified cases | Accuracy |
| --- | --- | --- |
| Varied threshold | 14 | 91.62% |
| Fixed threshold (0.5) | 16 | 90.42% |

From the table we can see that the method of optimising the threshold is slightly superior in that it allows us to classify two more fault cases. The different thresholds that were selected during the classification are shown in Figure 2 below.

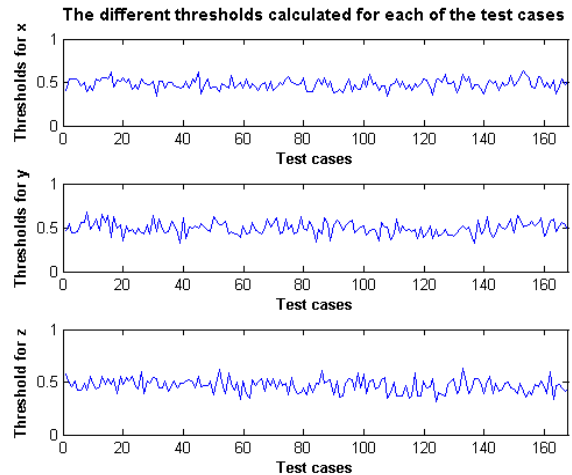

Figure 2: An illustration of the different thresholds that were selected for fault positions x, y and z.

The method used in this paper therefore gives a slightly better accuracy than the Bayesian trained neural network used by Marwala [2].

**CONCLUSION**

In this paper, pseudomodal properties obtained from the FRF have been calculated from vibration signal measured from a population of cylindrical shells. These properties have been used as inputs into a TS neuro-fuzzy model. A model selection process reveals that the optimum number of fuzzy rules for accurate classification is four as it allows for a high accuracy with a low variance. The TS neuro-fuzzy model classifies the faults in the cylindrical shells with an accuracy of 91.42%, which is an improvement over what the probabilistic neural network has been able to do in the past.

**ACKNOWLEDGMENTS**

The authors gratefully acknowledge the financial contributions and support obtained from the National Research Fund (NRF), Department of Science and Technology (DST) and the Central Building Research Institute (CBRI). It is through the funding of both this institutions that this project was possible.